\documentclass[journal]{IEEEtran}

\usepackage[utf8]{inputenc}
\usepackage[T1]{fontenc}

\usepackage{amsmath}
\usepackage{amsfonts}
\usepackage{amsopn}
\usepackage{nicefrac}

\usepackage{graphicx}

\DeclareGraphicsExtensions{.pdf,.png,.jpg,.eps}

\usepackage{booktabs}
\usepackage{array}
\usepackage{tabularx}
\usepackage{multirow}
\usepackage{threeparttable}
\usepackage{adjustbox}
\usepackage{subcaption}
\usepackage{algorithm}
\usepackage{algpseudocode}

\usepackage{microtype}
\usepackage{xcolor}
\usepackage{url}
\usepackage{enumitem}

\usepackage[
    breaklinks=true,
    colorlinks=true,
    citecolor=blue,
    linkcolor=blue,
    urlcolor=blue
]{hyperref}

\title{Beyond Static Costs: Learning-Dynamics Aware Loss Functions for Long-Tailed Classification}

\author{
Varad Shinde,
Nikhil Kumar Shrey,
Magesh Rajasekaran,
Md Saiful Islam Sajol,
Harshil Bhargava,
Subhajit Sidanta,
Supratik Mukhopadhyay,
and Yimin Zhu%
}

\begin{document}
\maketitle






\begin{abstract}
\textit{
Deep learning models in computer vision face significant challenges when trained on long-tailed datasets, where a few majority classes dominate while many minority classes are severely underrepresented. Such imbalances frequently arise in real-world scenarios such as rare species recognition, manufacturing fault detection, and medical image understanding, leading to biased models that underperform on tail classes. Existing reweighting methods typically rely on static class frequencies to penalize the model, ignoring the dynamic nature of how effectively a network actually learns a class over time. We address this by introducing a novel Learning-Dynamics Aware Loss (LDAL) function that shifts the focus from static sample counts to dynamic learning progress. LDAL framework adjusts class weights continuously by leveraging: (i) the strength of learned feature representations (semantic scale), (ii) the intrinsic learning difficulty of each class, measured via the Shannon entropy of its predictions, and (iii) an inter-epoch regularizer term that tracks prediction shifts between consecutive epochs to stabilize training and avoid local minima. LDAL is purely a objective function which incurs negligible computational overhead while adapting to the feature learning of the model. Experimental results on multiple benchmark datasets demonstrate that our approach significantly surpasses state-of-the-art reweighting loss functions, providing an optimal trade-off between accuracy and generalizability. The source code is available at \textcolor{blue}{https://github.com/sdm2026/ldal}}.
\end{abstract}

\section{Introduction.}
\label{sec:intro}

Developing reliable and accurate machine learning models requires the availability of balanced datasets with uniform class representation. However, in many real-world use cases, curating a perfectly balanced dataset that represents diverse deployment scenarios is nearly impossible. As a result, many real-world datasets exhibit varying degrees of imbalance, typically following a long-tailed distribution \cite{yang2022survey}. For example, in satellite image recognition in urban environments \cite{basu2015deepsat,collier2018progressively}, water-bodies may be underrepresented compared to buildings.  Long-tailed learning aims to improve inference accuracy for minority (tail) classes without sacrificing the performance of majority (head) classes, a challenge that has profound implications for practical computer vision applications \cite{roselli2019managing}. Examples include rare disease identification using medical imaging and rare species recognition in ecological monitoring, where accurate and unbiased predictions across all classes are critical.

According to the comprehensive taxonomy established by recent surveys in deep long-tailed learning \cite{zhang2023deeplongtailedlearningsurvey}, existing approaches for handling class imbalance fall primarily into three broad categories: (i) information augmentation (IA) \cite{kim2020m2m, park2022majority, wang2021rsgsimpleeffectivemodule, han2005borderline, zhong2021improving}, (ii) class re-balancing, which prominently includes class-sensitive learning \cite{lin2017focal, cui2019classbalancedlossbasedeffective, ren2020balancedmetasoftmaxlongtailedvisual, cao2019learningimbalanceddatasetslabeldistributionaware, park2021influencebalancedlossimbalancedvisual, legate2023reweightedsoftmaxcrossentropycontrol}, and (iii) module improvement (MI) \cite{dong2017class, huang2016learning, liu2019large, zhu2020inflated, ouyang2016factors, zhong2021improvingcalibrationlongtailedrecognition, zhou2020bbn, wang2020long, zhang2022self}. IA techniques attempt to balance distributions through data oversampling, undersampling, or feature transfer, but they often introduce new types of biases or substantially increase computational costs. MI approaches, such as decoupled training or ensemble learning, enhance generalization capability but typically require large-scale computation and complex architectural changes, limiting their practicality for existing pipelines. 

In contrast, class re-balancing methods specifically class-sensitive learning approaches that reweight or remargin the loss function penalize errors according to the class distribution. 
However, existing class-sensitive learning solutions typically depend on static weighting schemes derived from fixed dataset label frequencies. Because they rely on static priors, they do not adapt to the dynamic nature of model training. The intrinsic "complexity" of learning a specific class (i.e., its learning difficulty) changes continuously as the network optimizes. Static schemes do not capture these nuances, ultimately affecting the model's ability to generalize to difficult tail classes without overfitting.

We introduce a novel class-sensitive learning method: \textbf{Learning-Dynamics Aware Loss (LDAL)}. Instead of relying on static sample counts, the framework dynamically assigns class weights based on the strength of the features learned in each epoch and the relative learning difficulty of each class. Furthermore, LDAL introduces an inter-epoch regularizer term that links consecutive training epochs to prevent the model from prematurely converging to local minima dominated by head classes. Using a deeper understanding of the interplay between class representation strength and prediction uncertainty, LDAL aims to enhance both the robustness and generalization capabilities of models trained on heavily skewed datasets. 

Recent work \cite{ma2025pursuingbetterdecisionboundaries} has demonstrated that measuring category information amount and learning difficulty yields superior decision boundaries for long-tailed object detection via loss functions. The evaluation seeks to validate that a similar information-theoretic framework is exceptionally effective and establishes highly competitive baselines within the realm of long-tailed image classification.

The contributions of this paper are  as follows:
\begin{enumerate}
    \item \emph{Learning-Dynamics Aware Loss (LDAL)}: We propose a novel LDAL that dynamically adjusts class weights based on running prediction frequencies and entropy-based class difficulty, moving beyond traditional static weighting schemes. 
    \item \emph{Inter-Epoch Regularizer Term}: We introduce a lightweight penalty term that tracks prediction shifts between consecutive training epochs. This enforces stable optimization and prevents the model from settling into head-class-dominated local minima.
    \item \emph{Plug-and-play integration}: The LDAL framework operates purely at the objective level. It can be seamlessly integrated into existing machine learning architectures with negligible computational overhead, while pairing up with any other classification loss function.
    \item \emph{Extensive evaluation on long-tailed benchmarks}: We extensively experimented with benchmark datasets to demonstrate that LDAL significantly outperforms state-of-the-art class-sensitive loss functions. We also report the comparisons of LDAL with advance techniques, which uses heavy architectures for module improvements and data augmentation. The source code and datasets are available at \cite{anonymousiclrgit}.
\end{enumerate}
\section{Methodology.}
\label{sec:methodology}

Reweighting-based methods are highly convinient because they are easy to implement, require no complex architectural modules, and incur minimal training overhead compared to Information Augmentation (IA) or Module Improvement (MI) techniques. However, existing reweighting methods even those that incorporate label distribution priors directly into the logits, such as Balanced Softmax \cite{ren2020balancedmetasoftmaxlongtailedvisual} rely entirely on static dataset statistics. In reality, deep neural networks learn dynamically; the ``complexity'' of learning features for a particular class fluctuates significantly across epochs. We propose the Learning-Dynamics Aware Loss (LDAL) framework, which retains the simplicity of reweighting while dynamically adapting class weights based on real-time feature representation strength and intrinsic learning difficulty.

\subsection{Motivation: Beyond Static Label Frequencies.}
During training, a model encounters more samples from dominant (head) classes, leading to richer and stronger feature representations than those of tail classes. However, penalizing a model strictly based on inverse class frequency overlooks the fact that some classes are inherently Easier-to-Learn (ETL) while others are Difficult-to-Learn (DTL), regardless of their sample size. 

In the LDAL framework, the difficulty of learning a class is measured by the Shannon entropy of its post-softmax prediction probabilities across training batches. Confident predictions yield lower entropy, indicating an ETL class. Conversely, highly uncertain predictions yield higher entropy, indicating a DTL class. By shifting the penalty from raw sample counts to these dynamic learning metrics, LDAL ensures that the model does not over-penalize a tail class simply because it is rare, provided the network has already learned it effectively.

\subsection{The LDAL Formulation.}
The LDAL function acts as an auxiliary penalty that can be added to standard classification objectives. In our evaluation, we integrate it alongside the standard cross-entropy loss. For a batch of $B$ samples, the combined loss is:

\begin{equation}\label{eq:ldal_loss}
\resizebox{.9\linewidth}{!}{$
\begin{aligned}
\mathcal{L}
&= -\frac{1}{B}\sum_{b=1}^{B}\sum_{k=1}^C y^{(b)}_{\mathrm{true},k}\,\log p^{(b)}_{k} \\
&\quad + \frac{1}{C}\sum_{i=1}^C
\frac{\big(\gamma_i \tilde{N}_{i} + r_i\big)^2}
{\frac{1}{BC}\displaystyle \sum_{b=1}^B \sum_{k=1}^C 
\big(\mathrm{sg}(z^{(b)}_{k}) - (e_i)_{k}\big)^2 + \epsilon}
\end{aligned}
$}
\end{equation}

where:
\begin{itemize}
  \item $C$ is the total count of classes and $B$ is the batch size.
  \item $z^{(b)}\in\mathbb{R}^C$ denotes the model logits (pre-softmax) for sample $b$, with $z^{(b)}_k$ its $k$-th coordinate, and $p^{(b)}_k = \mathrm{softmax}(z^{(b)})_k$ the corresponding predicted probability for class $k$.
  \item $e_i\in\{0,1\}^C$ is the one-hot vector for class $i$, such that $(e_i)_k=\delta_{k,i}$.
  \item $\tilde{N}_{i} = \frac{C}{B}\sum_{b=1}^{B} p^{(b)}_i$ is the normalized differentiable soft prediction count for class $i$ in the current batch.  The $C/B$ factor normalizes for the number of classes, so that $\mathbb{E}[\tilde{N}_i] \approx 1$ under a uniform predictive distribution regardless of $C$. Using a continuous soft-count rather than a discrete $\mathrm{argmax}$ count maintains the computational graph, allowing the optimizer to backpropagate through the numerator and actively penalize the over-prediction of head classes during gradient descent.
  \item $\gamma_i$ is the dynamic importance weight for class $i$.
  \item $r_i$ is the inter-epoch regularizer term for class $i$.
  \item $\epsilon = 1.0$ is a stability constant added to prevent division by zero.
  \item $\mathrm{sg}(\cdot)$ denotes the stop-gradient operator.
\end{itemize}
\let\oldmathcal\mathcal
\renewcommand{\mathcal}[1]{%
  \ifstrequal{#1}{T}
    {\oldmathcal{U}}
    {\oldmathcal{#1}}%
}
\textbf{Denominator Design.} The denominator $\frac{1}{BC}\sum_{b,k}(\mathrm{sg}(z^{(b)}_k)-(e_i)_k)^2$ computes the mean squared deviation between the batch of model logits and the one-hot encoding of class $i$, averaged over all $B$ samples and $C$ dimensions. The $1/(BC)$ normalization converts the raw sum into a mean, keeping the denominator scale consistent across different batch sizes and class counts. This contributes as a continuous scaling factor that normalizes each class's penalty contribution based on how closely the model's current output distribution aligns with that class. The stop-gradient operator $\mathrm{sg}(\cdot)$ ensures this term functions purely as a scalar normalizer: without it, the optimizer could minimize the loss by inflating the denominator i.e., pushing logits away from all class prototypes which would degrade learned representations rather than improve classification.

Note that while $\tilde{N}_{i}$ in the numerator maintains gradient connectivity to enable optimization, the stop-gradient on the denominator creates an intentional asymmetry: the optimizer must resolve the penalty by genuinely adjusting its predicted class distribution, not by manipulating the normalizing factor.

\subsection{Dynamic Class Importance ($\gamma_i$).}
\label{sec:gamma}

Prior work \cite{ma2023delvingsemanticscaleimbalance} refers to the strength of feature representations as ``semantic values'', which are consistently larger for heavily exposed head classes. We define the semantic scale $S_i$ for class $i$ as the squared mean magnitude of its stored logit vectors:
\begin{equation}\label{eq:si}
S_i = \left(\frac{1}{|F_i|}\sum_{l=1}^{|F_i|}\|f_{il}\|\right)^2
\end{equation}
where $F_i$ is the set of logit vectors accumulated for class $i$ across training batches within the current epoch.

The learning difficulty of class $i$ is quantified by the mean Shannon entropy of its per-sample prediction distributions:
\begin{equation}\label{eq:entropy}
H_i = \frac{1}{|\mathcal{C}_i|}\sum_{b \in \mathcal{C}_i}\left(-\sum_{k=1}^{C} \mathrm{sg}(p^{(b)}_k) \log \mathrm{sg}(p^{(b)}_k)\right)
\end{equation}
where $\mathcal{C}_i = \{b : y^{(b)}_{\mathrm{true}} = i\}$ is the set of samples with true label~$i$. The stop-gradient on $p^{(b)}_k$ prevents the model from artificially manipulating entropy to reduce the loss.

The dynamic importance parameter $\gamma_i$ combines both signals:
\begin{equation}\label{eq:gamma}
\gamma_i = \min\!\left(\frac{S_i}{1 + \max_j(S_j) \cdot H_i},\;\tau\right)
\end{equation}
where $\tau = 5.0$ is an upper bound that keeps the importance weights within a stable range during training.

The intuition is straightforward: if a head class is predicted frequently ($\tilde{N}_{i}$ is high) and has strong semantic values with low entropy ($\gamma_i$ is high), the multiplicative penalty $\gamma_i \times \tilde{N}_{i}$ grows large. This forces the model to reduce its over-reliance on the head class, subsequently allowing it to focus on learning tail classes. 

To prevent catastrophic overfitting to the tail classes as their $\tilde{N}_{i}$ naturally rises, the $\gamma_i$ variable acts as a scaling factor. As the network successfully learns a tail class, its semantic scale $S_i$ increases and its entropy $H_i$ drops, naturally raising $\gamma_i$ and keeping the penalty term in check.

\subsection{Inter-Epoch Regularization ($r_i$).}
\label{sec:regularizer}

Training on highly imbalanced data often leads to abrupt parameter shifts between epochs, risking premature convergence to local minima dominated by head classes. LDAL counters this by introducing an inter-epoch regularizer $r_i$ for a designated set of target minority classes $\mathcal{T}$.

\let\oldhat\hat
\renewcommand{\hat}[1]{%
  \ifstrequal{#1}{N}
    {\oldmathcal{M}}
    {\oldmathcal{#1}}%
}

Let $\hat{N}^{(t)}_i$ denote the cumulative hard prediction count for class $i$ accumulated across batches processed so far within epoch $t$, and let $\hat{N}^{(t-1)}_i$ denote the finalized total count from the previous epoch. Note that $\hat{N}^{(t)}_i$ is computed via $\mathrm{argmax}$ and is distinct from the differentiable soft-count $\tilde{N}_i$ used in the loss numerator. The regularizer is defined as:
\begin{equation}\label{eq:ri}
r_i = \begin{cases}
+\alpha & \text{if } i \in \mathcal{T} \text{ and } \hat{N}^{(t)}_i < \hat{N}^{(t-1)}_i \\[4pt]
-\alpha & \text{if } i \in \mathcal{T} \text{ and } \hat{N}^{(t)}_i > \hat{N}^{(t-1)}_i \\[4pt]
\;\;0 & \text{otherwise}
\end{cases}
\end{equation}
where $\alpha = 1.0$ in all our experiments (see Section~\ref{sec:sensitivity} for the sensitivity analysis justifying this choice).

If the cumulative prediction count for a target minority class falls below its previous epoch's total, a positive penalty is applied, reinforcing the model's attention toward that class. Conversely, if predictions for that class have increased, a negative (reward) term is applied. For non-target classes, $r_i = 0$. All prediction statistics used in this computation are derived exclusively from training batches; no validation signals are used during the training loop.

\begin{figure*}[htbp]
\centering
\includegraphics[width=0.5\textwidth]{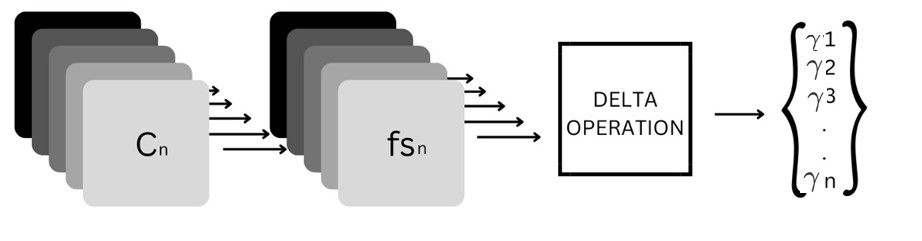} 
\vspace{-2mm}
\caption{Overview of the LDAL penalty computation. $C_n$ denotes class $n$, and $F_n$ represents its accumulated feature storage. The feature storage is passed through the Delta operation (Eq.~\ref{eq:gamma}), which combines the semantic scale $S_n$ with the class entropy $H_n$ to produce the dynamic importance weight $\gamma_n$.}
\vspace{-4mm}
\label{fig:trial2}
\end{figure*}

\section{Design Rationale and Training Procedure.}
\label{sec:solution-design}

\subsection{Empirical Motivation.}

Our hypothesis that classes should be penalized based on how well the model learns their features, rather than solely on sample count was motivated by examining the per-class classification report of ResNet-32 trained on CIFAR-10. We observed that classes such as \textit{Dog} and \textit{Cat} had noticeably lower accuracy compared to classes like \textit{Airplane} and \textit{Bus}, despite being trained under identical conditions and sometimes even having comparable sample counts. This observation suggested that intrinsic class difficulty plays a key role in how well a class's representation is learned, and that the loss function should account for this. As demonstrated in our ablation study (Section~\ref{sec:ablation-study}), penalizing classes based on their learning characteristics consistently outperforms static reweighting methods.

\subsection{The Role of Class Complexity.}
\label{sec:etl-dtl}

The distinction between Easy-to-Learn (ETL) and Difficult-to-Learn (DTL) classes is core point to the design of $\gamma_i$ (Eq.~\ref{eq:gamma}).

An ETL class has a compact feature distribution with low intra-class variance. For example, the \textit{Airplane} class in CIFAR-10 exhibits limited visual diversity. Most airplanes share a common shape making the class relatively straightforward to learn even with fewer samples. For such classes, the model produces confident predictions, yielding low class entropy $H_i$ (Eq.~\ref{eq:entropy}).

In contrast, a DTL class exhibits high intra-class variance. The \textit{Dog} class in CIFAR-10 encompasses an enormous variety of breeds, body shapes, and textures, making it inherently challenging. For DTL classes, the model's predictions are more uncertain, yielding high entropy $H_i$.

These observations lead to two important design principles:
\begin{enumerate}[leftmargin=*]
    \item ETL classes can be learned reliably even when they belong to the tail of the distribution. Therefore they should not receive heavy penalties just for being rare.
    \item DTL classes may remain poorly modeled despite having larger sample sizes. They require special monitoring during training.
\end{enumerate}

The $\gamma_i$ formulation (Eq.~\ref{eq:gamma}) tries to capture this: classes with higher entropy receive lower $\gamma_i$, ensuring that when their prediction count $\tilde{N}_i$ enters the loss, the resulting penalty is smaller. This in turn effectively protects difficult classes from excessive penalization. As training progresses and the model masters a class (entropy drops, semantic scale rises), its $\gamma_i$ naturally increases, dynamically shifting the optimizer's focus to the remaining hard classes. This self-balancing property eliminates the need for manual weight scheduling.

\subsection{Training Algorithm.}

\begin{algorithm}[H]
\caption{Training with LDAL}
\label{alg:ldal}
\begin{algorithmic}[1]
\Require Training data $D = (X, Y)$, number of classes $C$, epochs $T$, target minority classes $\mathcal{T}$, regularization strength $\alpha$
\Ensure Trained model parameters $w$
\State Initialize model weights $w$ randomly
\State $\hat{N}^{(0)}_i \leftarrow 0$ for all $i = 1, \ldots, C$ 
\For{epoch $t = 1$ \textbf{to} $T$}
    \State Reset: norm\_sum$_i$, norm\_count$_i$, ent\_sum$_i$, ent\_count$_i$, $\hat{N}^{(t)}_i \leftarrow 0 \;\; \forall i$
    \For{each mini-batch $(x, y)$ of size $B$ from $D$}
        \State $z \leftarrow f(x, w)$ 
        \State $p \leftarrow \mathrm{softmax}(z)$ 
        \State $\mathcal{L}_{\mathrm{CE}} \leftarrow -\frac{1}{B}\sum_{b} \sum_{k} y^{(b)}_k \log p^{(b)}_k$
        \State $\tilde{N}_i \leftarrow \frac{C}{B}\sum_{b} p^{(b)}_i \;\; \forall i$ 
        \State Update $\hat{N}^{(t)}_i$ via $\mathrm{argmax}$ counts 
        \State Accumulate per-class feature norms and entropy
        \State $S_i \leftarrow \left(\text{norm\_sum}_i \,/\, \text{norm\_count}_i\right)^2 \;\; \forall i$ 
        \State $H_i \leftarrow \text{ent\_sum}_i \,/\, \text{ent\_count}_i \;\; \forall i$
        \State $\gamma_i \leftarrow \min\!\big(S_i \,/\, (1 + \max_j S_j \cdot H_i),\;\tau\big) \;\; \forall i$
        \State $r_i \leftarrow \begin{cases} +\alpha & \text{if } i \in \mathcal{T},\; \hat{N}^{(t)}_i < \hat{N}^{(t-1)}_i \\ -\alpha & \text{if } i \in \mathcal{T},\; \hat{N}^{(t)}_i > \hat{N}^{(t-1)}_i \\ 0 & \text{otherwise} \end{cases}$
        \State $\mathcal{L}_{\mathrm{LDAL}} \leftarrow \frac{1}{C}\sum_{i} \frac{(\gamma_i \tilde{N}_i + r_i)^2}{\frac{1}{BC}\sum_{b,k}(\mathrm{sg}(z^{(b)}_k) - (e_i)_k)^2 + \epsilon}$ 
        \State $w \leftarrow w - \eta \nabla_w (\mathcal{L}_{\mathrm{CE}} + \mathcal{L}_{\mathrm{LDAL}})$ 
    \EndFor
\EndFor
\end{algorithmic}
\end{algorithm}

Algorithm~\ref{alg:ldal} summarizes the complete LDAL training procedure. At each epoch boundary, the running statistics (feature norms, entropy accumulators, prediction counts) are reset to avoid OOM errors. Within each mini-batch, the algorithm performs the following steps:
\begin{enumerate}[leftmargin=*, nosep]
    \item Computes the standard cross-entropy loss $\mathcal{L}_{\mathrm{CE}}$.
    \item Computes the differentiable soft prediction counts $\tilde{N}_i$ via softmax (maintaining the computational graph for backpropagation).
    \item Accumulates feature norms and prediction entropy statistics per class using the true labels.
    \item Derives the dynamic importance weights $\gamma_i$ from the accumulated semantic scales and entropies.
    \item Evaluates the inter-epoch regularizer $r_i$ for target minority classes $\mathcal{T}$ using cumulative hard prediction counts.
    \item Computes the LDAL penalty and updates model weights via backpropagation on $\mathcal{L}_{\mathrm{CE}} + \mathcal{L}_{\mathrm{LDAL}}$.
\end{enumerate}
All statistics including feature norms, prediction counts, and entropy values are computed exclusively from training batches. No validation data or external signals are used within the training loop.

Together, the dynamic $\gamma_i$ weights and the inter-epoch regularizer $r_i$ promote balanced learning: $\gamma_i$ steers the model's focus based on what has been learned so far, while $r_i$ penalizes stagnation on minority classes across epoch boundaries. Empirically, we observe that this combination yields stable optimization and smooth convergence across all evaluated benchmarks (Section~\ref{sec:evaluation}).

\section{Evaluation.} 
\label{sec:evaluation}
All experiments were conducted on an Intel Xeon server equipped with a 64-core CPU and a single Tesla V100-SXM2-32GB GPU. 
The implementation details, including hyperparameter settings for all datasets have been adopted from the most recent work from CSL category, AREA \cite{10378171}. 

\subsection{Baselines and Taxonomy.}
\label{sec:baselines}

For evaluation of Learning-Dynamics Aware Loss (LDAL) under fair comparisons, we organize our baselines using the deep long-tailed learning taxonomy established by Zhang et al. \cite{zhang2023deeplongtailedlearningsurvey}. Under this framework, various previous works are categorized by their core mechanisms: Class-Sensitive Learning (CSL), Decoupled Training (DT), Ensemble Learning, Representation Learning (RL), and Transfer Learning (TL). 

Since LDAL operates entirely as a standalone objective function requiring no multi-stage training or architectural modifications, it falls strictly within the \textbf{Class-Sensitive Learning (CSL)} category. Accordingly, our primary benchmarks are other pure CSL methods. These include all the proposed methods such as Cross-Entropy Loss (CE) \cite{zhang2018generalizedcrossentropyloss}, Focal Loss \cite{lin2018focallossdenseobject}, Class-Balanced Loss (CB) \cite{cui2019classbalancedlossbasedeffective}, and Influence Balanced Loss (IB) \cite{park2021influencebalancedlossimbalancedvisual} many more, as well as recent state-of-the-art objective formulations like LDAM \cite{cao2019learningimbalanceddatasetslabeldistributionaware} , SoLar \cite{wang2022solarsinkhornlabelrefinery} and AREA \cite{10378171}. 

For a complete perspective of the domain, we also provide separate reference tables detailing advanced, module-heavy architectures. These include decoupled frameworks like MisLAS \cite{zhong2021improvingcalibrationlongtailedrecognition} (DT), ensemble routing networks like RIDE \cite{wang2022longtailedrecognitionroutingdiverse} and SADE \cite{zhang2022selfsupervisedaggregationdiverseexperts}, prototype memory networks like Focal-SAM \cite{li2025focalsamfocalsharpnessawareminimization} (RL), and modern foundation models like LIFT \cite{shi2025liftlightweightfinetuninglongtail} (TL). We introduce this partition to clarify that while these methods achieve state-of-the-art absolute accuracy, direct numerical comparisons between a lightweight loss function and computationally intensive, multi-stage, or multi-expert architectures are fundamentally asymmetric.

\subsection{Results.}
\setcounter{footnote}{0}
\label{sec:results}

We evaluate LDAL across multiple scales of imbalance. For CIFAR-10, we experiment with imbalance ratios ($p$) of 50 and 100, and $p \in \{50, 100, 200\}$ for CIFAR-100. To assess scalability, we utilize ImageNet-LT (constructed via Pareto distribution \cite{liu2019largescalelongtailedrecognitionopen}) and iNaturalist-2018, a real-world dataset exhibiting severe natural long-tailed skew.

\subsubsection{CIFAR-10 and CIFAR-100 Evaluation.}
Table \ref{Table2_CSL} reports the top-1 accuracy for pure CSL methods using a ResNet-32 backbone, while Table \ref{Table3_Advanced} provides context against advanced approaches.

\begin{table}[ht]
\centering
\renewcommand{\arraystretch}{1.2}
\resizebox{\columnwidth}{!}{%
\begin{tabular}{@{}lccccc@{}}
\toprule
\multirow{2}{*}{\textbf{Pure CSL Methods}} & \multicolumn{2}{c}{\textbf{CIFAR-10}} & \multicolumn{3}{c}{\textbf{CIFAR-100}} \\ \cmidrule(lr){2-3} \cmidrule(l){4-6}
 & $p=100$ & $p=50$ & $p=200$ & $p=100$ & $p=50$ \\ \midrule
CE (Baseline) & 70.40 & 74.80 & 34.84 & 38.32 & 43.85 \\
CE + CB \cite{cui2019classbalancedlossbasedeffective} & 74.57 & 79.27 & 36.23 & 39.60 & 45.32 \\
Focal + CB \cite{cui2019classbalancedlossbasedeffective} & 74.60 & 79.30 & 35.62 & 39.60 & - \\
LDAM-DRW \cite{cao2019learningimbalanceddatasetslabeldistributionaware}& 77.00 & 80.90 & - & 42.04 & - \\
Focal ($\gamma=1$) \cite{lin2018focallossdenseobject} & 70.40 & 76.70 & 35.62 & 38.41 & 44.32 \\
Focal ($\gamma=2$) \cite{lin2018focallossdenseobject} & 69.59 & 76.52 & 34.75 & 38.39 & 43.70 \\
Focal ($\gamma=0.5$) \cite{lin2018focallossdenseobject} & 70.23 & 76.72 & 35.00 & 38.69 & 44.12 \\
LDAM \cite{cao2019learningimbalanceddatasetslabeldistributionaware} & 73.40 & 76.80 & - & 39.60 & - \\
Meta-Weight-Net \cite{NEURIPS2019_e58cc5ca} & 75.21 & 80.06 & 37.91 & 42.09 & 46.74 \\
IB \cite{park2021influencebalancedlossimbalancedvisual} & 78.26 & 81.70 & 37.31 & 42.14 & 46.22 \\
LDAM-DRW + SSP & 77.83 & 82.13 & - & 43.43 & - \\
AREA \cite{10378171} & 78.88 & 82.68 & 43.85 & 48.83 & 51.77 \\
SoLar ($\phi=0.3, 0.05$) \cite{wang2022solarsinkhornlabelrefinery} & 76.64 & 83.80 & - & - & 46.18 \\ \midrule
\textbf{LDAL (Ours)} & \textbf{80.19} & \textbf{84.41} & \textbf{45.04} & \textbf{49.79} & \textbf{52.72} \\ \bottomrule
\end{tabular}%
}
\caption{Comparison of pure CSL loss functions on CIFAR-10 and CIFAR-100 with various imbalance ratios ($p$). All methods operate purely at the objective level. LDAL results report the best top-1 accuracy across all runs.\protect\footnotemark[1]}
\label{Table2_CSL}
\end{table}

\begin{table}[!h]
\resizebox{\columnwidth}{!}{%
\begin{tabular}{@{}llccccc@{}}
\toprule
\multirow{2}{*}{\textbf{Methods}} & \multirow{2}{*}{\textbf{Category}} & \multicolumn{2}{c}{\textbf{CIFAR-10}} & \multicolumn{3}{c}{\textbf{CIFAR-100}} \\ \cmidrule(lr){3-4} \cmidrule(l){5-7}
 & & $p=100$ & $p=50$ & $p=200$ & $p=100$ & $p=50$ \\ \midrule
MisLAS \cite{zhong2021improvingcalibrationlongtailedrecognition} & DT & 82.10 & 85.70 & - & 47.00 & 52.30 \\
RIDE (3-experts) \cite{wang2022longtailedrecognitionroutingdiverse} & Ensemble & - & - & - & 48.60 & 51.40 \\
SADE \cite{zhang2022selfsupervisedaggregationdiverseexperts}& Ensemble & - & - & - & 49.80 & 53.90 \\
Focal-SAM \cite{li2025focalsamfocalsharpnessawareminimization} & RL & 77.20 & 82.20 & - & 44.00 & 48.10 \\
Balanced Softmax \cite{ren2020balancedmetasoftmaxlongtailedvisual} & Sampling+CSL & 84.90 & - & 45.50 & 50.80 & - \\ 
\textbf{LDAL (Ours)} & Pure CSL & 80.19 & 84.41 & 45.04 & 49.79 & 52.72 \\ \bottomrule
\end{tabular}%
}
\caption{Reference accuracies for advanced, module-heavy methods on CIFAR datasets. Categories denote Decoupled Training (DT), Ensemble Learning, Representation Learning (RL), and Transfer Learning (TL).\protect\footnotemark[1]}
\label{Table3_Advanced}
\end{table}

The results validate that LDAL dynamically rebalances training far more effectively than static priors. By adjusting penalties based on real-time feature strength and prediction entropy (as detailed in Section 3.3), LDAL avoids over-penalizing tail classes that are intrinsically "Easy-to-Learn" while maintaining necessary focus on "Difficult-to-Learn" classes. Compared to the baseline Cross-Entropy loss and its class-balanced variant (CE + CB) \cite{cui2019classbalancedlossbasedeffective}, LDAL yields improvements. Notably, LDAL consistently outperforms the recent state-of-the-art CSL method AREA \cite{10378171} across all imbalance ratios on CIFAR-100-LT, establishing a highly competitive new benchmark for pure loss functions.

\subsubsection{Large-Scale Evaluation: ImageNet-LT and iNaturalist-2018.}
Tables \ref{Table4_Large_CSL} and \ref{Table5_Large_Advanced} document our results on the large-scale datasets, again strictly separated by taxonomy. Training utilized a ResNet-50 backbone for 120 epochs on ImageNet-LT and 160 epochs on iNaturalist-2018.

\begin{table}[ht]
\centering
\renewcommand{\arraystretch}{1.2}
\resizebox{\columnwidth}{!}{%
\begin{tabular}{@{}lcc@{}}
\toprule
\textbf{Pure CSL Methods} & \textbf{ImageNet-LT} & \textbf{iNaturalist-2018} \\ \midrule
CE (Baseline) & 38.88 & 57.30 \\
Class-Balanced-Loss \cite{cui2019classbalancedlossbasedeffective}& 40.85 & 61.12 \\
Focal loss \cite{lin2018focallossdenseobject}& 30.50 & 58.03 \\
LDAM \cite{cao2019learningimbalanceddatasetslabeldistributionaware}& 41.86 & 64.58 \\
CE-DRW \cite{cao2019learningimbalanceddatasetslabeldistributionaware}& - & 63.73 \\
CE-DRS \cite{cao2019learningimbalanceddatasetslabeldistributionaware}& - & 63.56 \\
IB \cite{park2021influencebalancedlossimbalancedvisual}& - & 65.39 \\
FSR \cite{zhang2021learningfastsamplereweighting}& - & 65.52 \\
LDAM-DRW \cite{cao2019learningimbalanceddatasetslabeldistributionaware}& 45.74 & \textbf{68.00} \\
AREA \cite{10378171}& 49.53 & - \\ \midrule
\textbf{LDAL (Ours)} & \textbf{50.10} & 67.10 \\ \bottomrule
\end{tabular}%
}
\caption{Comparison of pure CSL functions on large-scale datasets using ResNet-50. LDAL results report the best top-1 accuracy.\protect\footnotemark[1]}
\label{Table4_Large_CSL}
\end{table}

\begin{table}[htbp]
\centering
\renewcommand{\arraystretch}{1.2}
\resizebox{\columnwidth}{!}{%
\begin{tabular}{@{}llcc@{}}
\toprule
\textbf{Methods} & \textbf{Category} & \textbf{ImageNet-LT} & \textbf{iNaturalist-2018} \\ \midrule
OLTR \cite{liu2019largescalelongtailedrecognitionopen} & RL & 40.36 & - \\
Balanced Softmax \cite{ren2020balancedmetasoftmaxlongtailedvisual} & Sampling+CSL & 41.80 & - \\
NCM \cite{kang2020decouplingrepresentationclassifierlongtailed}& DT & 44.30 & 63.10 \\
Decoupling \cite{kang2020decouplingrepresentationclassifierlongtailed}& DT & 47.30 & 67.60 \\
MisLAS \cite{zhong2021improvingcalibrationlongtailedrecognition} & DT & 52.70 & 70.70 \\
RIDE (3-experts) \cite{wang2022longtailedrecognitionroutingdiverse} & Ensemble & 54.40 & 71.80 \\
SADE \cite{zhang2022selfsupervisedaggregationdiverseexperts} & Ensemble & 58.80 & 72.90 \\
LIFT \cite{shi2025liftlightweightfinetuninglongtail} & Fine Tuning (CLIP)& 77.00 & 79.20 \\ 
\textbf{LDAL (Ours)} & Pure CSL & 50.10 & 67.10 \\ \bottomrule
\end{tabular}%
}
\caption{Reference accuracies for advanced, module-heavy methods on large-scale datasets. Categories denote Representation Learning (RL), Decoupled Training (DT), Ensemble Learning, and Transfer Learning (TL).\protect\footnotemark}
\label{Table5_Large_Advanced}
\end{table}

On ImageNet-LT, LDAL achieves a significant +11.22\% absolute accuracy improvement over the baseline Cross-Entropy loss. Crucially, it formally surpasses AREA (49.53\%), demonstrating that LDAL's dynamic, entropy-aware weighting scales effectively to large, structurally complex datasets without requiring hyperparameter retuning. 

\footnotetext{A dash ($-$) indicates that the referenced paper did not report the results in these settings.}

Similarly, on iNaturalist-2018, LDAL reaches 67.10\%, improving upon the CE baseline by +9.8\%. While LDAM-DRW \cite{cao2019learningimbalanceddatasetslabeldistributionaware} reports a marginally higher 68.00\%, it uses Deferred Reweighting (DRW) a specialized learning rate schedule deployed in the final epochs rather than standard SGD. When controlling for strictly default training conditions, LDAL remains highly competitive. While advanced architectures (Table \ref{Table5_Large_Advanced}) unsurprisingly post higher numbers through ensemble routing or representation decoupling, LDAL achieves robust performance purely at the loss level, eliminting the need for multi-stage pipelines or complex architectural overhead.

\begin{table}[!]
\centering
\renewcommand{\arraystretch}{1.2}
\resizebox{\columnwidth}{!}{%
\begin{tabular}{@{}lccc@{}}
\toprule
\textbf{Dataset} & \textbf{Mean Acc.(\%)} & \textbf{Std. Deviation(±)} & \textbf{95\% Confidence} \\ \midrule
CIFAR-10-LT       & 79.77 & 0.73 & [79.13, 80.41] \\
CIFAR-100-LT      & 48.88 & 0.86 & [48.12, 49.63] \\
ImageNet-LT       & 49.67 & 0.66 & [49.09, 50.24] \\
iNaturalist-2018  & 66.53 & 0.83 & [65.80, 67.25] \\ \bottomrule
\end{tabular}%
}
\caption{Mean accuracy, standard deviation, and 95\% confidence interval of LDAL across five independent runs with different random seeds. CIFAR-10 and CIFAR-100 are at imbalance factor 100.}
\label{ci}
\end{table}

Finally, to verify the stability of LDAL, particularly the stabilizing effect of the inter-epoch regularizer ($r_i$) discussed in Section 3.4 we conducted five independent test runs for each dataset. Table \ref{ci} details the mean accuracy, standard deviation and 95\% confidence intervals. The low variance confirms that preventing abrupt parameter shifts between epochs leads to smooth, reliable convergence across diverse data distributions.
\section{Ablation Study.}
\label{sec:ablation-study}
Hypothesis of proposed Learning-Dynamics Aware Loss (LDAL), is tested with a series of ablation studies that provide insights into the contribution of each of its core components. All experiments in this section were performed by training ResNet-32 on CIFAR-10 and CIFAR-100 with an imbalance ratio of 100, utilizing Stochastic Gradient Descent (SGD) with a momentum of 0.9. Using standard long-tailed learning protocols, the initial learning rate was assigned 0.1 and decayed by a factor of 0.01 at epochs 160 and 180. 

\subsection{Gradient Flow Verification.}
\label{sec:gradient-flow}
A critical requirement for end-to-end training is for the LDAL auxiliary term to produce non-zero gradients that flow to the backbone parameters. Because LDAL operates on \emph{soft predictions} sums of softmax probabilities rather than hard argmax counts. The computational graph remains connected. We empirically verify this by isolating the LDAL and CE gradient contributions on the output logits at four checkpoints during training.

\begin{table}[!]
\centering
\setlength{\tabcolsep}{3pt}
\small
\begin{tabular}{@{}l c c c c c@{}}
\toprule
 & \multicolumn{2}{c}{\textbf{Grad Norm}}& & & \\
\cmidrule(lr){2-3}
\textbf{Epoch} & $\|\nabla_{\text{LDAL}}\|$ & $\|\nabla_{\text{CE}}\|$ & \textbf{Ratio} & \textbf{NZ\%} & \textbf{Val Acc} \\
\midrule
\multicolumn{6}{@{}l}{\textit{CIFAR-10-LT ($\rho{=}100$, $C{=}10$)}} \\
\midrule
1   & 0.0121 & 0.1009 & 0.120 & 100.0 & 18.60 \\
25  & 0.0150 & 0.0889 & 0.169 & 98.8  & 33.52 \\
100 & 0.0078 & 0.0628 & 0.125 & 99.1  & 56.65 \\
200 & 0.0105 & 0.0446 & 0.235 & 98.9  & 80.19 \\
\midrule
\multicolumn{6}{@{}l}{\textit{CIFAR-100-LT ($\rho{=}100$, $C{=}100$)}} \\
\midrule
1   & 0.0000 & 0.1215 & 0.000 & 95.4  & 3.57  \\
25  & 0.0112 & 0.1057 & 0.106 & 97.9  & 21.64 \\
100 & 0.0082 & 0.0696 & 0.118 & 89.2  & 35.88 \\
200 & 0.0113 & 0.0445 & 0.254 & 88.3  & 49.79 \\
\bottomrule
\end{tabular}
\caption{\textbf{Gradient flow analysis.} Gradient norms of the LDAL and CE components measured on the output logits during training of ResNet-32. The LDAL term maintains non-zero gradients throughout and its relative contribution grows as training matures.}
\label{tab:gradient-flow}
\end{table}

Table~\ref{tab:gradient-flow} reports the gradient norms, the ratio $\|\nabla_{\text{LDAL}}\| / \|\nabla_{\text{CE}}\|$, and the percentage of non-zero elements in the LDAL Jacobian. We have two important observations. First, the LDAL term maintains near-complete gradient coverage ($>88\%$ non-zero) throughhout training for both CIFAR-10 and CIFAR-100, which confirms that the differentiable soft-count formulation preserves the gradient flow. Second, the LDAL-to-CE gradient ratio generally increases from ${\sim}0.10$ to ${\sim}0.25$ over 200 epochs. So as the model's cross-entropy loss decreases which can be visualized as the basic features being learned. Meanwhile the relative influence of LDAL naturally grows, providing a stronger class-rebalancing signal especially in later epochs when model focuses on fine grained features.

\subsection{Component Ablation.}
\label{sec:component-ablation}

To quantify the contribution of each LDAL component, we train four ablated variants alongside the full model: (i) CE-only baseline, (ii) LDAL without entropy in $\gamma_i$, (iii) LDAL without the inter-epoch regularizer ($r_i{=}0$), and (iv) LDAL without semantic scale in $\gamma_i$.

\begin{table}[H]
\centering
\small
\setlength{\tabcolsep}{6pt}
\renewcommand{\arraystretch}{1.05}
\begin{tabular}{@{}lcc@{}}
\toprule
\textbf{Configuration} & \textbf{CIFAR-10} & \textbf{CIFAR-100} \\ \midrule
CE only (baseline) & 71.32 & 40.36 \\
LDAL w/o entropy & 76.15 & 44.84 \\
LDAL w/o semantic scale & 76.27 & 44.55 \\
LDAL w/o regularizer ($r_i$) & 78.73 & 45.80 \\
\textbf{LDAL (full)} & \textbf{80.19} & \textbf{49.79} \\ \bottomrule
\end{tabular}
\caption{Component ablation on CIFAR-10-LT and CIFAR-100-LT ($\rho{=}100$). Each row removes one component from the full LDAL.}
\label{tab:component-ablation}
\end{table}
\vspace{-2mm}

The variant with all components of LDAL outperforms every ablated variant on both datasets, confirming that all three components contribute meaningfully. Removing entropy or semantic scale causes comparable drops (${\sim}4\%$ on CIFAR-10, ${\sim}5\%$ on CIFAR-100), indicating that both signals are equally important in computing effective class weights. Remooving the regularizer ($r_i$) yields a smaller but consistent drop ($1.5$--$4\%$) showing that $r_i$ primarily benefits late-stage convergence. Even the weakest LDAL variant surpasses the CE baseline by $+4.2\%$ on CIFAR-100, demonstrating that our method provides substantial gains.

\subsection{Effectiveness and Sensitivity of Inter-Epoch Regularization ($r_i$).}
\label{sec:sensitivity}

To evaluate the inter-epoch regularization term ($r_i$), we conduct two complementary experiments: (i) a direct comparison of training with the regularizer active ($\alpha=1.0$) versus disabled ($\alpha=0$), and (ii) a sensitivity sweep over $\alpha \in \{0.5, 1.0, 2.0, 5.0, 10.0, 20.0\}$.

\subsubsection{Validation Curve Analysis.}

\begin{figure}[H]
    \centering
    \begin{subfigure}[b]{0.49\columnwidth}
        \centering
        \includegraphics[width=\textwidth]{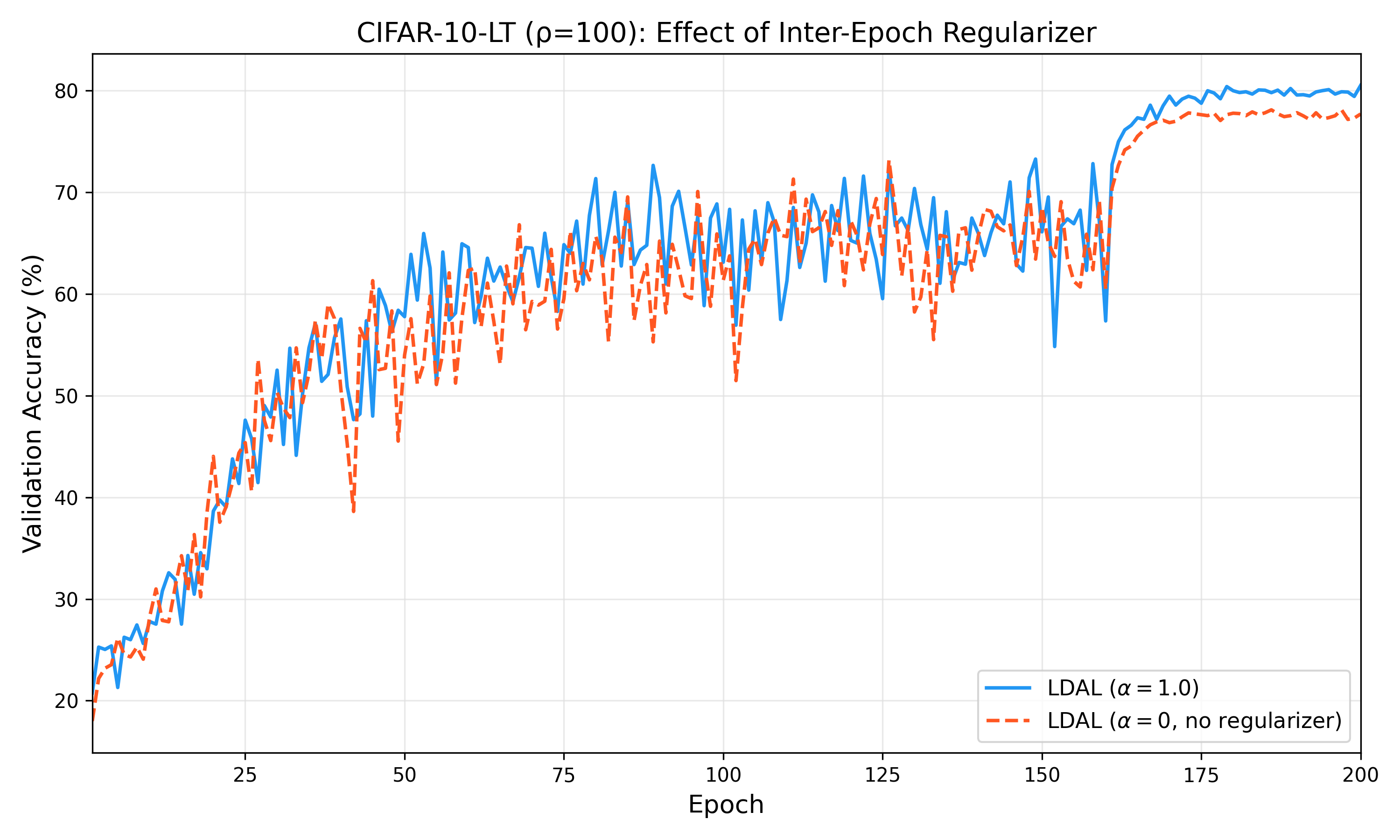}
        \caption{CIFAR-10-LT ($\rho{=}100$)}
        \label{fig:val_curve_c10}
    \end{subfigure}
    \hfill
    \begin{subfigure}[b]{0.49\columnwidth}
        \centering
        \includegraphics[width=\textwidth]{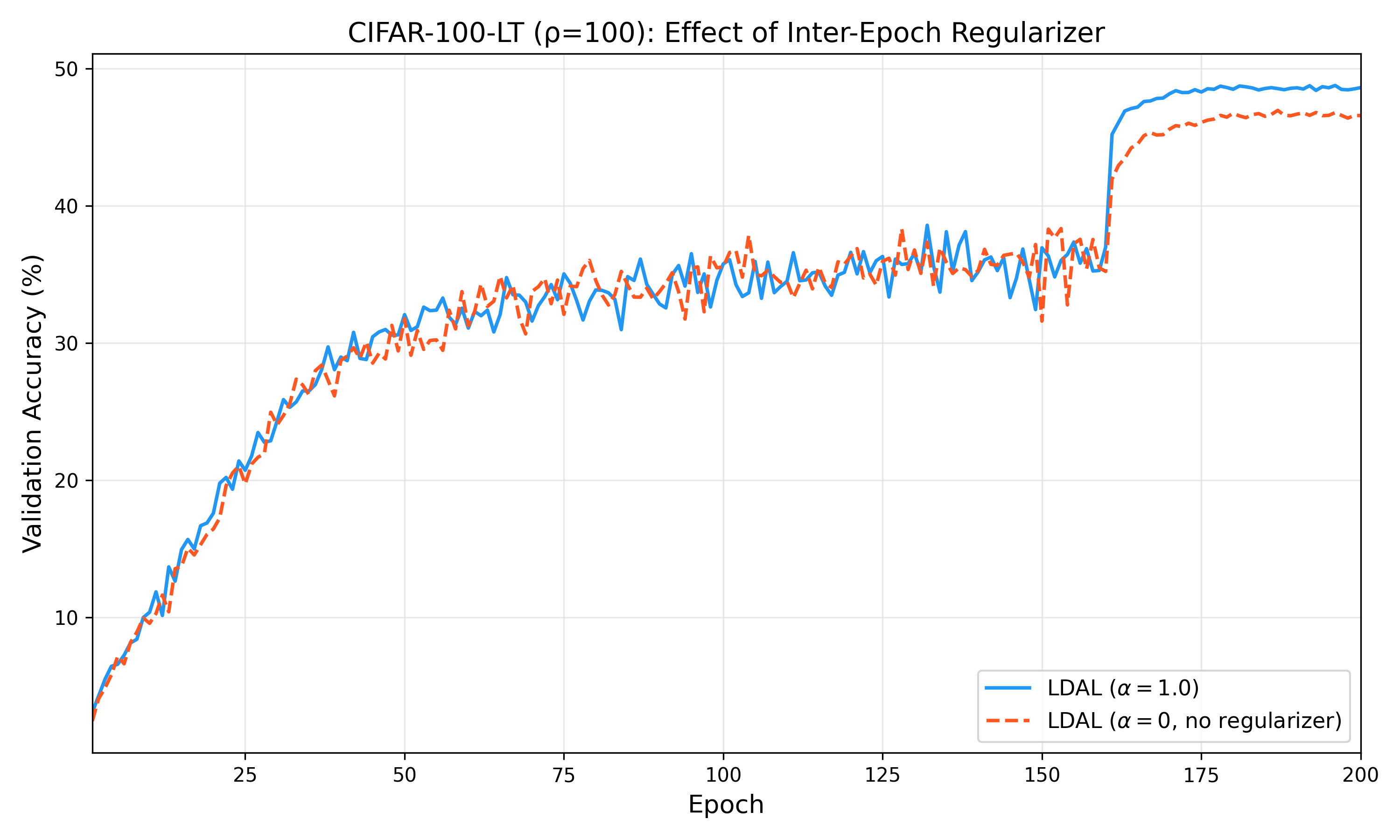}
        \caption{CIFAR-100-LT ($\rho{=}100$)}
        \label{fig:val_curve_c100}
    \end{subfigure}
    \caption{Validation accuracy curves comparing LDAL with $\alpha=1.0$ (solid blue) against $\alpha=0$ (dashed orange, no inter-epoch regularizer).}
    \label{fig:val_curves_both}
\end{figure}

Figure~\ref{fig:val_curves_both} overlays the epoch-wise validation accuracy for LDAL with and without the regularizer on both CIFAR-10-LT and CIFAR-100-LT.

During epochs 1--160, both configurations follow similar trajectories, indicating that $\gamma_i$ and cross-entropy dominate early optimization. The critical divergence appears after the learning rate decay at epoch 160: the regularized model ($\alpha{=}1.0$) converges to a higher accuracy plateau on both CIFAR-10-LT (${\sim}80\%$ vs.\ ${\sim}77\%$) and CIFAR-100-LT (${\sim}49\%$ vs.\ ${\sim}46\%$). Without the regularizer, the model settles into a lower local minimum dominated by head classes, confirming that $r_i$ has role in the late-stage convergence.

\textbf{Sensitivity to $\alpha$.}
To further characterize the regularizer, we sweep $\alpha \in \{0.5, 1.0, 2.0, 5.0, 10.0, 20.0\}$ on both datasets.

\begin{table}[H]
\centering
\small
\setlength{\tabcolsep}{6pt}
\renewcommand{\arraystretch}{1.05}
\begin{tabular}{@{}ccc@{}}
\toprule
$\alpha$ & \textbf{CIFAR-10} & \textbf{CIFAR-100} \\ \midrule
0.5 & 79.24 & 45.80 \\
\textbf{1.0} & \textbf{80.19} & \textbf{49.79} \\
2.0 & 78.50 & 46.48 \\
5.0 & 74.94 & 47.59 \\
10.0 & 53.17 & 44.63 \\
20.0 & 33.17 & 42.67 \\ \bottomrule
\end{tabular}
\caption{Sensitivity of LDAL to the reg strength $\alpha$.}
\label{tab:sensitivity}
\end{table}

As shown in Table~\ref{tab:sensitivity}, $\alpha{=}1.0$ yields the best accuracy on both datasets, confirming that the optimal regularization strength transfers across scales without dataset-specific tuning. For moderate values ($\alpha \in [0.5, 2.0]$), CIFAR-10-LT accuracy stays within $1.69\%$, while beyond $\alpha{=}5.0$ performance degrades sharply as regularization terms overwhelm the cross-entropy signal. Notably, CIFAR-100-LT is far more robust ($\Delta{=}7.12\%$ vs.\ $\Delta{=}47.02\%$ on CIFAR-10-LT), since the $1/C$ normalization naturally considers each class's contribution when $C{=}100$. Though we believe stronger analysis for optimum value of $\alpha$ is dependent on the datasets because of role of C in the loss function.

\subsection{Representing Class Difficulty with Entropy.}
\label{sec:entropy-proxy}

To validate that Shannon entropy is a meaningful measure of intrinsic class difficulty, we compute the Spearman rank correlation between per-class entropy $H_i$ recorded during training and the final per-class test accuracy on CIFAR-10-LT ($\rho{=}50$).

\begin{figure}[H]
    \centering
    \includegraphics[width=0.8\columnwidth]{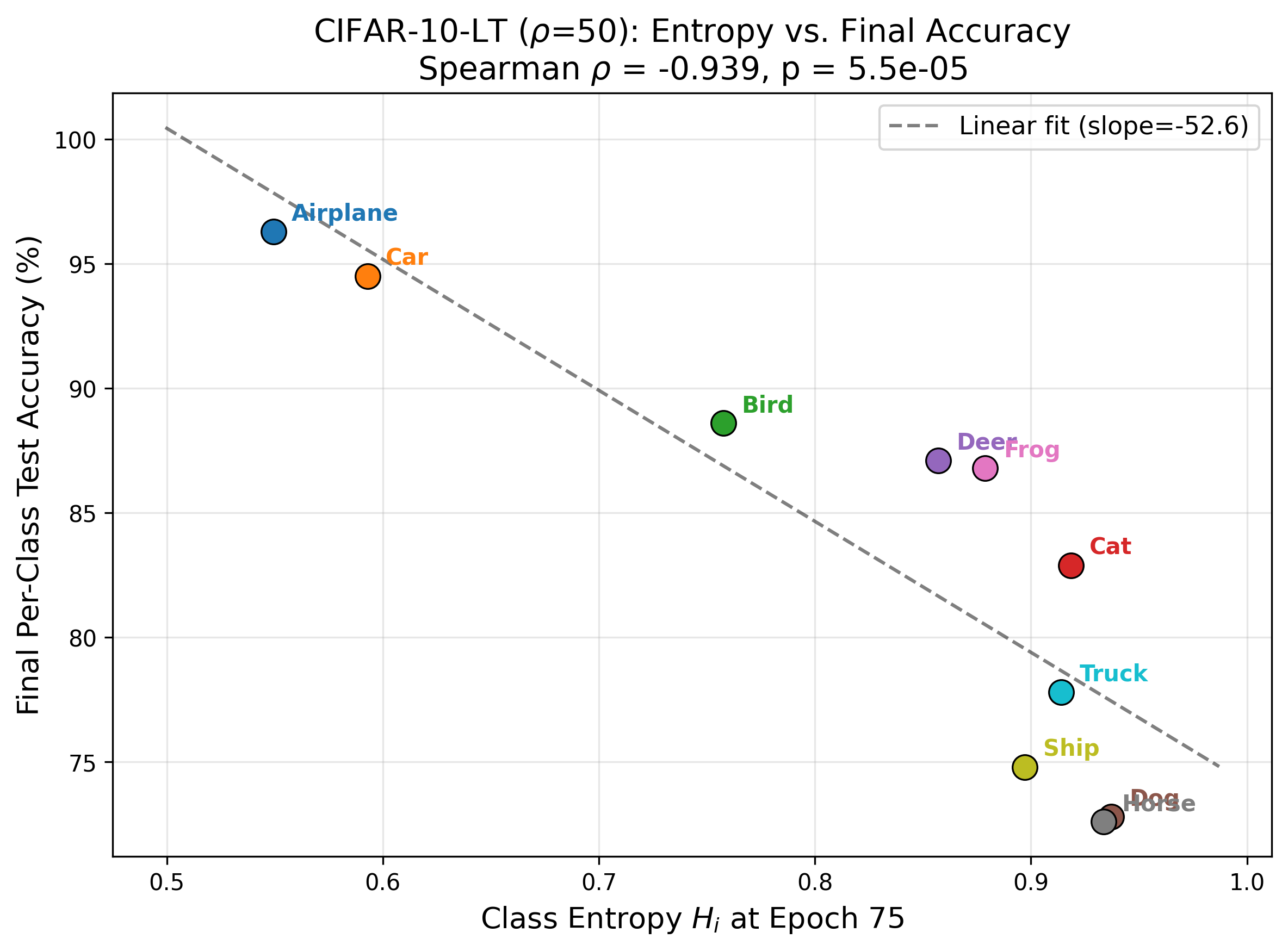}
    \caption{Per-class entropy $H_i$ at epoch 75 vs.\ final test accuracy on CIFAR-10-LT ($\rho{=}50$). Each point represents one class. Spearman $\rho = -0.94$, $p < 0.001$.}
    \label{fig:entropy-corr}
\end{figure}

Figure~\ref{fig:entropy-corr} shows a strong negative correlation (Spearman $\rho = -0.94$, $p < 10^{-4}$): classes with higher prediction entropy during training consistently achieve lower final accuracy. Notably, visually simple classes such as \textit{Airplane} and \textit{Car} exhibit low entropy and high accuracy, while semantically complex classes like \textit{Dog} and \textit{Horse} exhibit high entropy and low accuracy---independent of their position in the long-tailed distribution. This confirms that entropy captures intrinsic learning difficulty rather than merely reflecting sample count, justifying its use in the $\gamma_i$ formulation (Eq.~\ref{eq:gamma}).

\section{Conclusions and Future Work.}
\label{sec:conclusion}
This work introduces Learning-Dynamics Aware Loss (LDAL), a dynamic loss function that overcomes the limitations of static reweighting by utilizing semantic scale and prediction entropy to measure intrinsic class difficulty. The mechanism for shifting penalization away from raw sample counts and toward learning progress helps the framework to adapt intelligently. Using these components, LDAL achieves highly competitive top-1 accuracies without requiring complex multi-stage pipelines or architectural overhead. In future, we aim to extend the dynamic weighting mechanism to object detection and integrate transformer mechanisms instead of convolution backbones. The ultimate goal will be to develop adaptable and reliable real world services \cite{yau2005adaptable,yau2007automated} that do not suffer from bias due to imbalance in the training data.


    \bibliographystyle{plain}
    \bibliography{ldal_arxiv_ref}

\end{document}